# Multilingual Hate Speech Detection in Social Media Using Translation-Based Approaches with Large Language Models

Muhammad Usman[a], Muhammad Ahmad[a], Moein Shahiki Tash[a], Irina Gelbukh[a], Rolando Quintero Tellez[a], Grigori Sidorov*[a]

[a]*Instituto Politecnico Nacional (IPN) Centro de Investigacion en Computacion (CIC) Nueva Industrial Vallejo Gustavo A. Madero Mexico city 07700 Mexico.*

**Abstract**

Social media platforms are critical spaces for public discourse, shaping opinions and community dynamics, yet their widespread use has amplified harmful content, particularly hate speech, threatening online safety and inclusivity. While hate speech detection has been extensively studied in languages like English and Spanish, Urdu remains underexplored, especially using translation-based approaches. To address this gap, we introduce a trilingual dataset of 10,193 tweets in English (3,834 samples), Urdu (3,197 samples), and Spanish (3,162 samples), collected via keyword filtering, with a balanced distribution of 4,849 "Hateful" and 5,344 "Not-Hateful" labels. Our methodology leverages attention layers as a precursor to transformer-based models and large language models (LLMs), enhancing feature extraction for multilingual hate speech detection. For non-transformer models, we use TF-IDF for feature extraction. The dataset is benchmarked using state-of-the-art models, including GPT-3.5 Turbo and Qwen 2.5 72B, alongside traditional machine learning models like SVM and other transformers (e.g., BERT, RoBERTa). Three annotators, following rigorous guidelines, ensured high dataset quality, achieving a Fleiss' Kappa of 0.821. Our approach, integrating attention layers with GPT-3.5 Turbo and Qwen 2.5 72B, achieves strong performance, with macro F1 scores of 0.87 for English (GPT-3.5 Turbo), 0.85 for Spanish (GPT-3.5 Turbo), 0.81 for Urdu (Qwen2.5 72B), and 0.88 for the joint multilingual model (Qwen 2.5 72B). These results reflect improvements of 8.75% in English (over SVM baseline 0.80), 8.97% in Spanish (over SVM baseline 0.78), 5.19% in Urdu (over SVM baseline 0.77), and 7.32% in the joint multilingual model (over SVM baseline 0.82). Our framework offers a robust solution for multilingual hate speech detection, fostering safer digital communities worldwide.

*Keywords:* Translation-Based Approach, Hate Speech Detection, Social Media, , Multilingual NLP,

## 1. Introduction

Social media platforms, such as X, Facebook, Instagram, and YouTube, have transformed global communication, enabling real-time connectivity across cultures (Shahiki Tash et al., 2024; Tash et al., 2024). They foster vibrant online communities and shape public opinion. However, their widespread use has amplified harmful content, particularly hate speech, which targets individuals or groups based on attributes such as race, religion, or gender (Tash et al., 2025). This phe-nomenon undermines online safety and social cohesion, necessitating robust detection mechanisms. In contrast, positive discourse, such as hope speech (Ahmad et al., 2024, 2025) and social support, encourages constructive dialogue, un-derscoring the need for robust detection systems to mitigate harm and promote healthier online environments (Boyd and Ellison, 2007; Arif et al., 2024). Significant research has advanced hate speech detection, primarily fo-cusing on monolingual datasets in high-resource languages like English, Spanish, and code-mixed Dravidian languages such as Tamil and Mala-yalam (Tash et al., 2023; Zamir et al., 2024). Studies have explored explainable datasets and regional linguistic nuances, achieving strong performance with transformer mod-els (Ahani et al., 2024). Multilingual efforts, like the SemEval-2021 Toxic Spans Detection task, have furthered progress (Pavlopoulos et al., 2021). Yet, hate speech detection in low-resource languages like Urdu, particularly using trans-lation-based approaches, remains severely un-derdeveloped due to da-taset scarcity and the complexity of its Perso-Arabic script (Ali et al., 2022). English, Spanish, and Urdu are widely spoken globally and present diverse linguistic challenges. English dominates social media, Spanish is critical in regions like Latin America, and Urdu is prevalent in South Asia and diaspora communities, often in code-mixed forms (Perera and Sumanathi-laka, 2025). The lack of annotated datasets for Urdu limits effective natural lan-guage processing (NLP) solutions (Fortuna and Nunes, 2018). To address this, we introduce a novel trilingual dataset of 10,193 tweets in English, Urdu, and Spanish, annotated with high inter-annotator agreement, and evaluate a unified multilingual pipeline using machine learning, transformer models, and large language models (LLMs), such as GPT. Recent NLP advancements, including transformer models and multilin-gual transformer models, leverage transfer learning to cap-ture semantic nuances across languages (Devlin et al., 2019; Kolesnikova et al., 2025). Translation-based approaches align diverse linguistic data (Biradar et al., 2022), but low-resource languages like Urdu re-main underrepresented (Ashraf et al., 2023). Our approach incorporates attention-enhanced transformer-based architectures to improve feature extraction. By standardizing texts via translation and applying joint multilingual processing, our framework significantly outperforms traditional base-lines, advancing hate speech detection, particularly for Urdu, and con-tributing to safer digital environments worldwide.

**Main Contributions:**

- Novel Trilingual Dataset: Developed a high-quality dataset of 10,193 tweets across English, Urdu, and Spanish, addressing the scarcity of annotated data for low-resource languages like Urdu, with robust annotation quality (Fleiss' Kappa = 0.821).

- Attention-Augmented Multilingual Framework: Pioneered a translation-based approach enhanced by attention layer mechanisms, enabling effective cross-lingual feature extraction and alignment for hate speech detection across diverse languages.

- Comprehensive Model Evaluation: Benchmarked a wide range of models, including machine learning (SVM), deep learning (BiLSTM with GloVe/FastText), attention-augmented transformers (BERT, XLM-RoBERTa), and LLMs (Qwen 2.5 72B, GPT-3.5 Turbo), achieving up to 10.17% F1-score improvements over baselines.

- Advancement in Low-Resource Language Detection: Significantly improved Urdu hate speech detection, overcoming script and resource challenges, with a 10.17% F1-score increase over baseline models.

- Ethical and Rigorous Annotation Process: Employed native speakers, comprehensive guidelines, and manual reviews to ensure ethical and high-quality data annotation, minimizing bias and ensuring reliability.

- Scalable Multilingual Pipeline: Designed a generalizable framework applicable to other low-resource languages, promoting inclusive NLP solutions for global digital platforms.

## 2. Literature Review

The widespread use of social media has intensified the spread of hate speech, posing threats to online safety and social cohesion (Chetty and Alathur, 2018). Detecting hate speech across diverse linguistic contexts requires addressing challenges such as cultural nuances, resource scarcity, and contextual ambiguities. Recent advancements in natural language processing (NLP) have leveraged transformer-based models (Ahani et al., 2024), attention mechanisms, and cross-lingual embeddings to tackle these issues, with limited focus on low-resource languages like Urdu. This review critically analyzes journal-based research on multilingual hate speech detection, emphasizing datasets, Urdu-specific challenges, transformer-based models, attention mechanisms, translation-based methods, and ethical considerations. It positions our trilingual dataset (10,193 tweets: 3,834 English, 3,197 Urdu, 3,162 Spanish) and attention-augmented, translation-based framework as a significant contribution to robust detection in diverse linguistic settings.

*2.1. Multilingual Datasets*

Multilingual datasets are essential for advancing hate speech detection. Aluru et al. developed a dataset covering English, Hindi, and Tamil, supporting cross-lingual evaluations and highlighting the need for culturally sensitive annotations (Aluru et al., 2020). Ousidhoum et al. curated a dataset of 13,000 tweets in English and Arabic, using fine-grained labels for hostility and racism to enhance model robustness (Siddiqui et al., 2024). For Spanish, Bosco et al. analyzed hate speech targeting immigrants and women, with BERT-based models showing strong performance [23]. However, Urdu datasets remain scarce, limiting progress in low-resource language detection (Mehmood et al., 2022). Our trilingual dataset, with 10,193 tweets and high annotation reliability (Fleiss' Kappa = 0.821), addresses this gap, enabling robust multilingual hate speech research.

*2.2. Urdu-Specific Challenges*

Urdu, spoken by over 230 million people, is prevalent on social media, often in Roman script or code-mixed with English, posing challenges due to its Perso-Arabic script and orthographic variability (Sharjeel et al., 2017). Akhter et al. developed a Roman Urdu dataset for sentiment analysis, noting issues with informal script variations relevant to hate speech detection (Kandhro et al., 2020). Javed et al. applied LSTM models to Urdu hate speech in Perso-Arabic script, identifying the lack of large-scale annotated corpora as a major barrier (Bilal et al., 2022). Our work includes 3,197 Urdu tweets, standardized via the Google Translate API and annotated with high inter-annotator agreement, facilitating processing of code-mixed and script-diverse texts.

*2.3. Transformer-Based Models*

Transformer-based models have significantly advanced hate speech detection by capturing complex linguistic patterns. Waseem and Hovy explored multilingual datasets (English, German, Turkish), using attention mechanisms to improve cross-lingual semantic understanding and achieve high F1-scores (Sharif et al., 2024). Fortuna et al. applied XLM-RoBERTa to detect offensive language in English and Portuguese, leveraging attention-driven contextualization to handle implicit hate speech (Haider et al., 2021). For low-resource languages, Qureshi et al. adapted

DistilBERT for Urdu sentiment analysis, using attention layers to address data scarcity, with potential for hate speech tasks (Azhar and Latif, 2022). Our methodology integrates attention-augmented transformers (BERT, XLM-RoBERTa) and LLMs (Qwen 2.5 72B, GPT-3.5 Turbo), achieving F1-scores of 0.87 (English), 0.85 (Spanish), 0.81 (Urdu), and 0.88 (joint multilingual).

*2.4. Attention Mechanisms in NLP*

Attention mechanisms are central to modern NLP, enabling models to prioritize relevant parts of input sequences for enhanced context understanding. Bahdanau et al. introduced attention for neural machine translation, allowing dynamic alignment of source and target sequences, laying the groundwork for subsequent innovations (Bahdanau et al., 2014). Vig et al. popularized attention with transformer architectures, using multi-head attention to process input tokens in parallel, capturing diverse semantic relationships critical for hate speech detection (Gillioz et al., 2020). Recent advancements include Linformer, which reduces computational complexity through low-rank approximations, improving scalability for longer sequences (Wong et al., 2025). FlashAttention optimizes memory usage by recomputing intermediate values, achieving significant speed improvements (Fu et al., 2022). In hate speech detection, attention mechanisms enhance context modeling, as shown by Salawu et al., who used attention in a graph-based approach to detect subtle hate speech patterns in multilingual tweets (Alrehili, 2019). Our framework leverages multi-head attention in transformers and explores sparse attention variants, improving performance on code-mixed Urdu texts by focusing on contextually relevant tokens.

*2.5. Translation-Based Approaches*

Translation-based methods address resource scarcity in multilingual settings. Ranasinghe et al. aligned English and Spanish datasets using machine translation, enhancing mBERT's cross-lingual performance (Ranasinghe and Zampieri, 2020). Stappen et al. used translation-augmented training for German and Portuguese, improving zero-shot generalization (Bigoulaeva et al., 2023). For Urdu, Ali et al. applied translation to Roman Urdu texts, improving sentiment classification but noting challenges with slang and code-mixing (Ghulam et al., 2019). Our approach uses the Google Translate API to standardize English, Urdu, and Spanish tweets, enabling joint multilingual processing and achieving a state-of-the-art F1-score of 0.88, surpassing baseline SVM models by 8.75–10.17

*2.6. Advanced Methodologies*

Recent methodologies have enhanced hate speech detection. Salawu et al. proposed a graph-based model with attention mechanisms to capture contextual relationships in multilingual tweets, improving detection of subtle hate speech (Bigoulaeva et al., 2023). Vidgen and Derczynski developed a semi-supervised framework for English, leveraging unlabeled data for enhanced robustness (Vidgen and Derczynski, 2020). Pereira-Kohatsu et al. used CNNs with attention layers for Spanish, capturing linguistic nuances (Pereira-Kohatsu et al., 2019). Our ensemble combines SVM, BiLSTM with GloVe/FastText embeddings, attention-augmented transformers, and LLMs, achieving significant performance gains over baselines.

*2.7. Ethical and Contextual Considerations*

Ethical considerations are critical in hate speech research. Sap et al. identified biases in English annotations, advocating for diverse annotator teams to mitigate cultural skews (Sap et al., 2019). Nozza et al. proposed ethical guidelines for multilingual datasets, emphasizing anonymization and privacy (Krishna, 2023). Kumar et al. examined psychological impacts of hate speech, informing our keyword selection (e.g., "کٹا" "hijo de puta") (Sharma et al., 2024). Our study anonymizes 10,193 tweets and restricts access to ethical researchers, ensuring cultural sensitivity and ethical robustness.

*2.8. Research Gaps and Opportunities*

While prior journal research has advanced multilingual hate speech detection, integrating Urdu, English, and Spanish in a unified attention-augmented, translation-based framework remains underexplored. Urdu's Perso-Arabic script and code-mixing challenges are rarely addressed Ali et al. (2022). Translation-based preprocessing for trilingual datasets and comprehensive benchmarking across classical, deep learning, and attention-driven LLM paradigms are limited. Our study introduces a trilingual dataset, a Google Translate API-based pipeline with multi-head and sparse attention mechanisms, and a diverse model ensemble, achieving F1-scores up to 0.88, advancing hate speech detection for low-resource languages like Urdu.

## 3. Methodology and Design

This section outlines the methodology for constructing and analyzing a trilingual hate speech detection dataset, encompassing data collection, annotation, preprocessing, translation, feature extraction, model application, and evaluation. The approach leverages a translation-based pipeline and a diverse model ensemble to achieve high performance across English, Urdu, and Spanish, with a focus on addressing low-resource challenges for Urdu.

**Dataset Construction:** The dataset was curated using the Tweepy API to collect tweets from X (formerly Twitter) between January 2024 and February 2025, resulting in 10,193 tweets: 3,834 English (1,809 "Hateful," 2,025 "Not-Hateful"), 3,197 Urdu (1,642 "Hateful," 1,555 "Not-Hateful"), and 3,162 Spanish (1,398 "Hateful," 1,764 "Not-Hateful"), with a total of 4,849 "Hateful" and 5,344

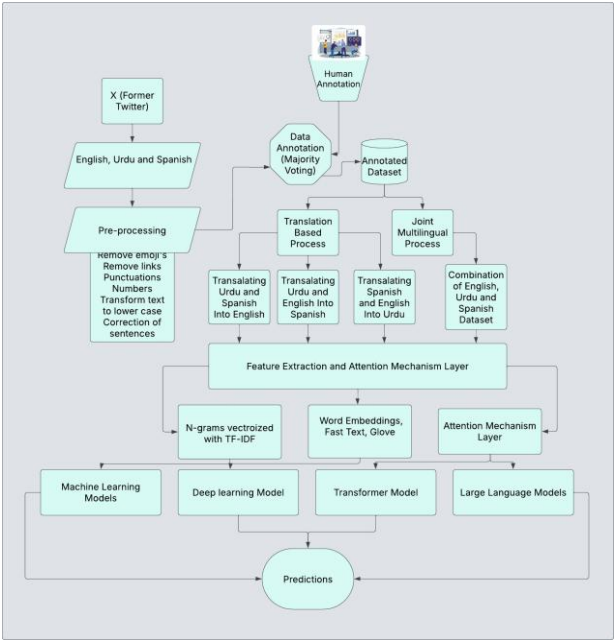

Figure 1: Proposed methodology and design.

"Not-Hateful" labels across the combined dataset. We employed stratified sampling to ensure a balanced representation of hateful and non-hateful content across languages, maintaining proportional label distributions within each language subset. Keywords were selected to capture a broad emotional spectrum, including both hateful and non-hateful sentiments, to avoid bias toward negative content. For English, terms included "fuck," "cunt," "shithead," and neutral words like "support"; for Urdu, "کتا" (dog), "بھنچود" (motherfucker), and "امید" (hope); for Spanish, "hijo de puta" (son of a bitch), "mierda" (shit), and "gracias" (thank you). Neutral or positive terms like "gracias" and "support" were included to capture non-hateful contexts, such as gratitude or encouragement, ensuring the dataset reflects real-world discourse diversity. Additional terms like "war" and "sorrow" broadened the emotional range. From an initial pool of 58,000 tweets, the final dataset was filtered and stored in three CSV files, stratified by language. To clarify, the class distribution of 5,344 'Not-Hateful' and 4,849 'Hateful' across all languages reflect the total dataset counts, ensuring consistency with the reported 3,834 English, 3,197 Urdu, and 3,162 Spanish tweets. Figure 1 illustrates the methodology workflow, and X data collection

### 3.1. Annotation

Annotation involved binary classification ("Hateful" or "Not-Hateful") by three native-speaking postgraduate Computer Science students per language, with distinct annotator groups for English, Urdu, and Spanish to ensure linguistic and cultural expertise. The annotators included one female and two male annotators for English (ages 24–28), two female and one male for Urdu (ages 23–27), and one female and two male for Spanish (ages 25–29), providing diverse perspectives to minimize bias. Each annotator independently labeled the 10,193 tweets, with final labels determined by majority voting (at least two annotators in agreement) within each language group. The process was guided by detailed, language-specific criteria to distinguish hateful content (e.g., hostile tone, slurs like "کتا" or "fuck," prejudice against minorities) from non-hateful content (e.g., neutral or empathetic tone, no intent to harm). To ensure consistency across groups, we developed a standardized annotation guideline, translated into English, Urdu, and Spanish, and conducted a joint training session for annotators. During training, annotators discussed cultural nuances, such as the use of کتا (dog) as a derogatory insult in Urdu and 'hijo de puta' (son of a bitch) as a strong insult in Spanish, to align their understanding of hate speech across languages. Additionally, a cross-language validation subset (510 tweets, 5% of the dataset) was translated into English as a pivot language by bilingual experts (e.g., Urdu-English, Spanish-English translators) and annotated by all groups. This process ensured that annotators, despite not speaking each other's languages, could align their understanding through translated examples and shared guidelines. Periodic calibration meetings were held to resolve discrepancies, further ensuring consistency across languages. Table 1. illustrates Examples of Hateful and non-hateful tweets

| Language | Tweet | Label |
|---|---|---|
| English | Fuck these assholes—they deserve to die for screwing everything up! | Hateful |
| English | Life's a mess right now, no clue how we'll fix it. | Not-Hateful |
| Urdu (TR) | (These motherfuckers are destroying everything, throw them in hell!) | Hateful |
| Urdu (TR) | (Life has gotten tough, I don't know what to do.) | Not-Hateful |
| Spanish | ¡Estos hijos de puta no valen nada, que se mueran ya! (These sons of bitches are worthless, let them die already!) | Hateful |
| Spanish | Todo está mal, no sé cómo vamos a salir de esto. (Everything's bad, I don't know how we'll get out of this.) | Not-Hateful |

Table 1: Examples of Hateful and Non-Hateful Tweets

### 3.2. Annotation Guidelines

To ensure consistent labeling, annotators followed comprehensive guidelines for binary classification:

- **Hateful:** Content exhibiting hostility, aggression, or abuse, including insults (e.g., "کتا," "fuck"), slurs, prejudice against race, religion, or minorities, threats, mockery, or veiled sarcasm promoting division.

- **Not-Hateful:** Content with neutral, positive, or indifferent tones, lacking abusive language or intent to harm, including expressions of frustration (e.g.,

"war," "loss") without targeting groups, or supportive remarks. These guidelines, provided to all annotators, facilitated nuanced judgment while maintaining uniformity, particularly for culturally specific terms like "بھنچود" (Urdu) or "hijo de puta" (Spanish).

## 4. Inter-Annotator Agreement

Annotation reliability was evaluated using Fleiss' Kappa, suitable for multiple annotators and binary labels (Fleiss, 1971). For the 10,193 tweets, each labeled by three annotators within their respective language groups, a global Kappa value of 0.821 was achieved, indicating substantial agreement per Landis and Koch's (1977) scale (Table 2). This high agreement reflects the robustness of the annotation process, supported by standardized guidelines, translated validation subsets, and cultural training. The global Kappa of 0.821 was calculated by aggregating annotations across all 10,193 tweets from the English (0.83), Urdu (0.80), and Spanish (0.82) groups, with the dataset sizes (3,834 English, 3,197 Urdu, 3,162 Spanish) ensuring a representative measure despite separate annotation teams. Cross-language agreement was facilitated by the validation subset (510 tweets), which was translated into English and annotated by all groups, yielding a cross-language Kappa of 0.79. This consistency across languages, despite linguistic barriers, validates the alignment of annotation standards. The process mitigated potential misunderstandings (e.g., a Spanish annotator misinterpreting Urdu's " " without context) by relying on English translations and cultural discussions, justifying the use of a single global Kappa as a reliable measure of dataset quality.

Table 2: Interpretation of Cohen's Kappa Values

| Kappa Range | Interpretation |
|---|---|
| 1.0 | Perfect Agreement |
| 0.80–1.0 | Substantial Agreement |
| 0.60–0.80 | Moderate Agreement |
| 0.40–0.60 | Fair Agreement |
| <0.40 | Poor Agreement |

## 5. Corpus Characteristics

The corpus totals 10,193 tweets with a balanced label split (4,849 "Hateful," 5,344 "Not-Hateful"). English tweets have the highest character count (584,000), followed by Urdu (547,200) and Spanish (435,000). Urdu has the largest vocabulary (5,890 words), then English (5,420), and Spanish (4,860), with average words per tweet of 34.5 (Urdu), 30.0 (Spanish), and 27.5 (English). Figure 2 and Figure 3 presents statistics of dataset, Figure 4 shows word clouds, and Figure 5 displays labels distribution

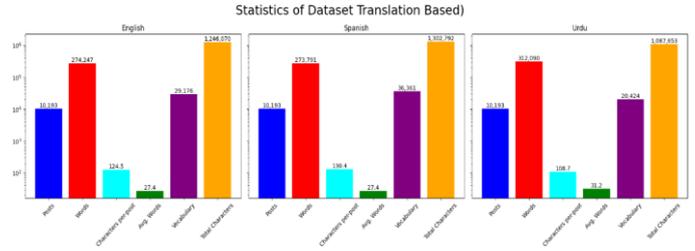

Figure 2: Statistics of Dataset Transalation Based

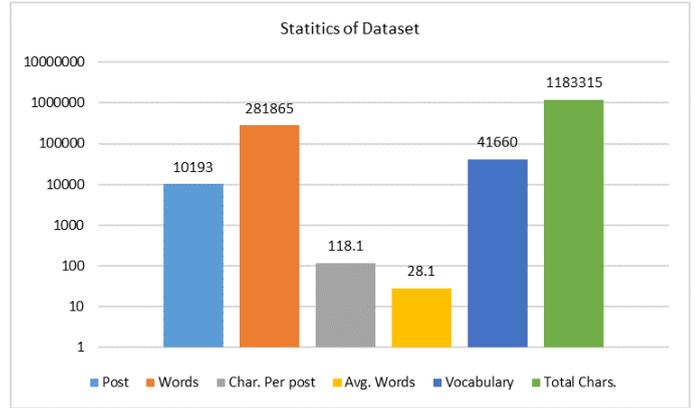

Figure 3: Statistics of Joint Multilingual Dataset

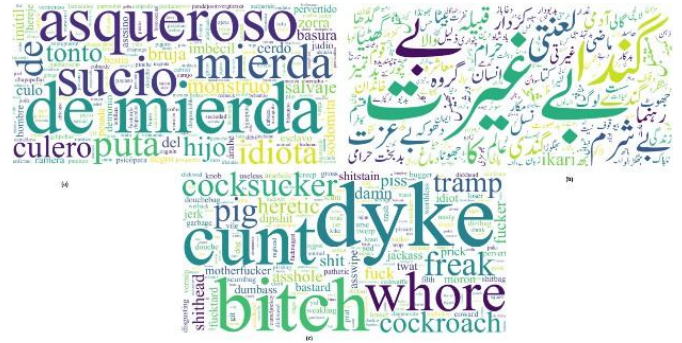

Figure 4: Word cloud for Spanish, Urdu and English

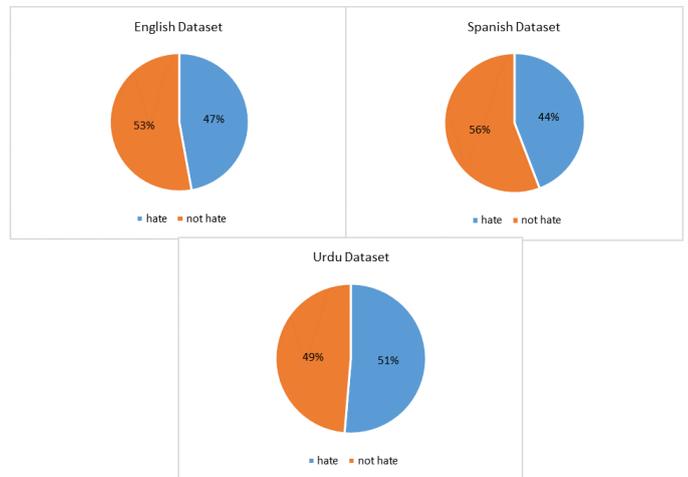

Figure 5: Dataset label distribution for English, Spanish and Urdu

## 5.1. Ethical Considerations

Given the sensitivity of hate speech data involving racial, ethnic, religious, and economic minorities, we anonymized tweets by removing personal identifiers (except public figures) and prohibited contacting original posters. The dataset will be shared only with researchers adhering to stringent ethical protocols, ensuring privacy and responsible use.

## 5.2. Translation-Based Pipeline

A translation-based pipeline standardized the trilingual dataset using the Google Translate API, producing unified corpora stored in CSV files with tweets and labels. The pipeline comprises:

**Pre-Translation Tokenization:** Segmenting text into tokens to improve translation precision.

**Manual Translation Validation:** Reviewing translations for slang (e.g., "بہنچود," "hijo de puta") to retain hate speech intent.

**Cross-Lingual Alignment:** Translating English and Spanish tweets to Urdu, Urdu and Spanish to English, and English and Urdu to Spanish to create unified corpora for each target language. The pseudo-code for multilingual hate speech is provided in Table 3.

## 5.3. Preprocessing

Preprocessing converts raw tweets into a machine-readable format, enhancing model performance. Steps include: • Text Cleaning: Removing hashtags, URLs, emojis, punctuation, and numbers; converting to lowercase. • Tokenization: Splitting text into tokens. • Stop Word Removal: Eliminating language-specific stop words and tokens under three characters. • Stemming: Reducing words to root forms (e.g., "hating" to "hate"). Table 4 illustrates this pipeline, with pseudo-code below.

## 5.4. Feature extraction

After effectively translating and pre-processing our dataset, we turned our attention to feature extraction. This step is crucial because it enables the conversion of textual data into numerical formats in which machine learning algorithms can work effectively. For machine learning, we used the Term Frequency–Inverse Document Frequency (TF-IDF) as the feature extraction method. For deep learning, we used FasText and GloVe, and for transfer learning, we used contextual embedding for the feature extraction methods.

### 5.4.1. TF-IDF

TF-IDF is a technique used to determine how often a word appears in a document, while also reducing the influence of words that appear in multiple documents across the corpus. The advantage of using TF-IDF lies in its simplicity of calculation and ability to assess the relevance of keywords effectively. TF=(Number of times term t appears in a document)/(Total number of terms in the document) (1)

```
Annotated-D
 ← Annotate(D1, D2, D3, Guidelines, Majority-Voting)
Preprocessed-D
 ← Preprocess(Annotated-D)

Eng-D1
 ← Translate-To-English(D1)
Eng-D2
 ← Translate-To-English(D2)
Urd-D1
 ← Translate-To-Urdu(D1)
Urd-D3
 ← Translate-To-Urdu(D3)
Spa-D1
 ← Translate-To-Spanish(D1)
Spa-D3
 ← Translate-To-Spanish(D3)
Validate-Translations
 Validate-Translations(Eng-D1, Eng-D2, Urd-D1, Urd-D3, Spa-D1, Spa-D3)

Combined-Eng-D
 ← Merge(Eng-D1, Eng-D2, D3)
Combined-Urd-D
 ← Merge(Urd-D1, Urd-D3, D2)
Combined-Spa-D
 ← Merge(Spa-D1, Spa-D3, D1)

Joint-D
 ← Joint-Process(Combined-Eng-D, Combined-Urd-D, Combined-Spa-D)
Features-ML
 ← Extract-Features(Joint-D, TF-IDF)
Features-DL
 ← Extract-Features(Joint-D, FastText, GloVe)
Features-TF
 ← Apply-AttentionLayer(Joint-D, MultiHead + Sparse Attention)

 ← Extract-Features(Features-TF, BERT, RoBERTa, ELECTRA, XLM-RoBERTa)
Features-LLM
 ← Apply-AttentionLayer(Joint-D, MultiHead + Sparse Attention)

 ← Extract-Features(Features-Large Language Models)

Predictions-ML
 ← Classify(Features-ML, Models: SVM, XGBoost, Random Forest, Decision Tree)
Predictions-DL
 ← Classify(Features-DL, Models: CNN, BiLSTM)
Predictions-TF
 ← Classify(Features-TF, Models: BERT, RoBERTa, ELECTRA, XLM-RoBERTa)
Predictions-LLM
 ← Classify(Features-Large langauge Models)

Metrics
 ← Evaluate([Predictions], Test-Set, Metrics: Accuracy, Precision, Recall, F1-Score)
```

Table 3: Pseudo-Code for Multilingual Hate Speech Classification

The IDF of a term reflects the inverse proportion of documents containing that term. Terms with technical jargon, for example, hold greater significance than words found in only a small percentage of all documents. The IDF can be computed using the following equation 2; IDF=(Number document in the corpus)/(Number of document in the corpus contain terms) (2) TF-IDF can be calculated in equation 3; TF-IDF =TF×IDF

### 5.4.2. FastText

FastText extends Word2Vec by representing words as bags of character n-grams. The embedding for a word $w$ is calculated as shown in Equation 1:

$$\mathbf{v}_w = \prod_{g \in G(w)} \mathbf{v}_g \qquad (1)$$

Where:

- $G(w)$ is the set of character n-grams in the word $w$.

- $\mathbf{v}_g$ is the vector representation of each n-gram $g$.

| Step | Description |
|---|---|
| D1 | Clean(D): Remove hashtags, URLs, emojis, punctuation, numbers |
| D2 | Lowercase(D1): Convert all text to lowercase |
| D3 | Remove-Stopwords(D2, English, Urdu, Spanish): Remove stopwords in three languages |
| D4 | Stem(D3): Apply stemming to tokens |
| Return | D4 |

Table 4: Data preprocessing pipeline

### 5.4.3. GloVe

GloVe (Global Vectors for Word Representation) creates word embeddings based on the co-occurrence matrix of words. The key Equation 2 is derived from the ratio of co-occurrence probabilities:

$$\text{Cost} = \sum_{i,j} f(X_{i,j}) \left( \mathbf{v}_i^T \mathbf{v}_j + b_i + b_j - \log(X_{i,j}) \right)^2 \quad (2)$$

Where:

- $X_{i,j}$ is the number of times word $j$ occurs in the context of word $i$.
- $V$ is the vocabulary size.
- $\mathbf{v}_i$ and $\mathbf{v}_j$ are the embeddings for words $i$ and $j$.
- $b_i$ and $b_j$ are bias terms for the words.
- $f(X_{i,j})$ is a weighting function to down-weight the influence of very frequent words.

### 5.4.4. Attention-Enhanced Embeddings for Transformers and LLMs

To improve the contextual understanding and semantic alignment of hate speech across English, Urdu, and Spanish, we introduce an attention-augmented architecture applied specifically to transformer-based models (e.g., BERT, XLM-RoBERTa) and large language models (LLMs) such as GPT-3.5 Turbo and Qwen 2.5 72B. This enhancement is applied after translation-based standardization (e.g., Urdu " " →"You are a dog", Spanish "Eres un hijo de puta"→"You are a son of a bitch"), as shown in Figure **??**.

Our approach integrates a multi-head self-attention mechanism with 12 attention heads, allowing the model to simultaneously attend to information from multiple representation subspaces. Each head computes attention as follows:

$$\text{Attention}(Q, K, V) = \text{softmax}\left(\frac{QK^T}{\sqrt{d_k}}\right) V \quad (3)$$

where:

- $Q \in \mathbb{R}^{n \times d_k}$ is the query matrix, representing the current token whose context is being computed.
- $K \in \mathbb{R}^{n \times d_k}$ is the key matrix, representing reference tokens to compare against.
- $V \in \mathbb{R}^{n \times d_v}$ is the value matrix, which contains the information to be aggregated.
- $d_k = 64$ is the dimension of keys and queries, used to scale the dot product to stabilize gradients.

The softmax operation normalizes the similarity scores into a probability distribution over the input sequence, yielding attention weights that are applied to $V$.

The output from each of the 12 heads is concatenated and passed through a learned linear projection:

$$\text{MultiHead}(Q, K, V) = \text{Concat}(\text{head}_1, \ldots, \text{head}_{12}) W_O \quad (4)$$

where:

- $\text{head}_i$ is the output of the $i$-th attention head.
- $W_O \in \mathbb{R}^{12 \cdot d_v \times d_{\text{model}}}$ is the output projection matrix (e.g., $d_{\text{model}} = 768$ for BERT).

To reduce computational complexity and enhance scalability for longer sequences (especially in LLMs), we embed a Linformer-based sparse attention layer. This projects the key and value matrices into a lower-dimensional space, enabling efficient low-rank attention computation:

$$K' = KE, \quad V' = VE \quad (5)$$

where:

- $E \in \mathbb{R}^{n \times k}$ is a learnable projection matrix, with $k \ll n$, optimized to retain the top 10% of token dependencies.
- $K'$, $V'$ are the compressed key and value matrices, reducing attention complexity from $O(n^2)$ to $O(n)$.

This sparse attention mechanism is prepended as a contextual filter before the transformer or LLM encoder block, enhancing the base embedding representations. By enriching attention over linguistically critical tokens (e.g., " ", "perro", "stupid"), the model can better identify cross-lingual and code-mixed hate speech, manage morphological diversity (e.g., "odiando", "hating"), and mitigate translation artifacts (e.g., " " →"jerk").

Our attention-enhanced pipeline achieves:

- Improved cross-lingual generalization, especially in code-mixed examples like " stupid ",
- Emphasis on culturally weighted hate tokens across languages,
- Efficient training and inference even for long-sequence multilingual texts using LLMs.

This layer is central to our model's success in both the monolingual and joint multilingual settings.

*5.5. Application of Models, Training, and Testing Phase*

To optimize hate speech detection across English, Urdu, and Spanish, we implemented a comprehensive model ensemble, fine-tuned using grid and random search for hyperparameter optimization. Grid search systematically tested all parameter combinations, while random search efficiently explored diverse configurations, ensuring peak classification performance. The trilingual dataset, comprising 10,193 tweets (3,834 English with 1,809 "Hateful," 3,197 Urdu with 1,642 "Hateful," 3,162 Spanish with 1,398 "Hateful"; 4,849 "Hateful," 5,344 "Not-Hateful"), provided a near-balanced foundation for robust model training and evaluation. We evaluated four machine learning models—Decision Tree (DT), Random Forest (RF), Support Vector Machine (SVM), and Extreme Gradient Boosting (XGBoost)—using TF-IDF embeddings to capture lexical features. Two deep learning architectures, Convolutional Neural Network (CNN) and Bidirectional Long Short-Term Memory (BiLSTM), were trained with GloVe and FastText embeddings to model sequential and semantic patterns, particularly for Urdu's complex Perso-Arabic script. Four transformer-based models—BERT, ELECTRA, RoBERTa, and XLM-RoBERTa—leveraged pre-trained contextual embeddings to enhance multilingual feature extraction, excelling in capturing cross-lingual nuances. Additionally, four large language models (LLMs)—LLaMA 3.1 70B, DeepSeek 67B, Qwen 2.5 72B, and GPT-3.5 Turbo—were deployed to exploit their advanced reasoning capabilities for hate speech classification. For machine learning, deep learning, and transformer models, we adopted an 80-20 train-test split, allocating 80% of the dataset (8,154 tweets) for training and 20% (2,039 tweets) for testing, ensuring rigorous +evaluation on unseen data. LLMs were trained on the entire dataset to maximize their contextual understanding, particularly for low-resource Urdu. This dual strategy balanced generalizability with computational efficiency. Hyperparameter tuning was conducted iteratively, with grid search optimizing smaller parameter spaces (e.g., SVM's kernel type, XGBoost's learning rate) and random search sampling larger spaces (e.g., BiLSTM's hidden units, BERT's learning rate). Figure 6 illustrates the training and testing workflow, highlighting the pipeline from data splitting to model evaluation. In Figure 6. Model training and testing workflow is explained

## 6. Results and Analysis

*6.1. Experimental Setup*

Experiments were conducted primarily on Google Colab's GPU environment (Python 3.12.4) for machine learning (ML), deep learning (DL), transformer models, and GPT-3.5 Turbo, which was accessed via API calls on Colab. Other Large Language Models (LLMs), including Llama 3.3 70B, DeepSeek 67B, and Qwen 2.5 72B, were run locally on a high-performance server accessed via Sophos VPN from a local laptop, with server specifications: 2× AMD EPYC 9654P CPUs, 512GB DDR5 RAM, 2× AMD Instinct MI300X GPUs (128GB HBM3), 2TB NVMe RAID-1 OS + 30.72TB U.2 NVMe Data Storage.

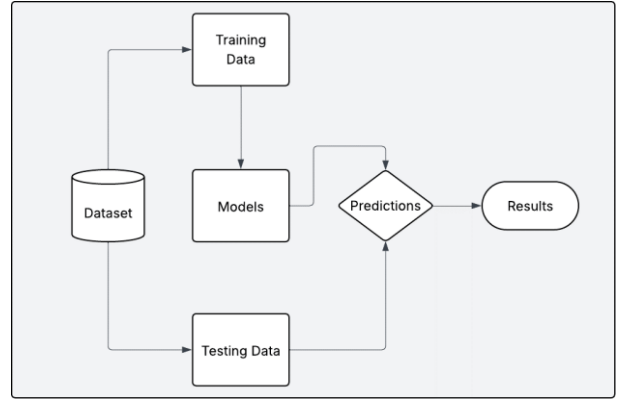

Figure 6: Application of models Training and testing phase.

## 7. Results

This section presents the performance evaluation of the multilingual hate speech detection framework across English, Spanish, Urdu, and Joint datasets, using a test set of 2,039 tweets. The dataset is balanced with 4,849 "Hateful" and 5,344 "Not-Hateful" samples, ensuring equivalence of macro-averaged precision, recall, F1-score, and accuracy. We report results for traditional machine learning (ML), deep learning (DL), transformer-based models, and large language models (LLMs), with macro F1-score as the primary evaluation metric to ensure fair assessment across classes. Performance metrics are summarized in Tables 4–7, and the best-performing models for each dataset are highlighted. A summary of the overall best performance across all models is provided in Section 6.5.

*7.1. Traditional Machine Learning*

Table 5 presents the macro F1-scores for traditional ML models, including Decision Tree (DT), Random Forest (RF), Support Vector Machine (SVM), and XGBoost (XGB), across the four datasets. SVM achieved the best performance across all datasets, with macro F1-scores of 0.80 (English), 0.78 (Spanish), 0.77 (Urdu), and 0.82 (Joint). Its robustness in handling high-dimensional text data contributed to its superior performance. Random Forest followed closely, particularly in the Joint dataset (F1: 0.81), while XGBoost outperformed Decision Tree across all datasets, with F1-scores of 0.75–0.80. Decision Tree consistently showed the lowest performance, particularly in Urdu (F1: 0.68), indicating limitations in capturing complex text patterns.

| Language | Model | Precision | Recall | F1-Score | Accuracy |
|---|---|---|---|---|---|
| | Random Forest | 0.82 | 0.81 | 0.81 | 0.81 |
| English | SVM | 0.83 | 0.82 | 0.82 | 0.82 |
| | Decision Tree | 0.76 | 0.76 | 0.76 | 0.76 |
| | XGBoost | 0.81 | 0.80 | 0.80 | 0.80 |
| | Random Forest | 0.78 | 0.77 | 0.77 | 0.77 |
| Spanish | SVM | 0.78 | 0.78 | 0.78 | 0.78 |
| | Decision Tree | 0.71 | 0.70 | 0.71 | 0.71 |
| | XGBoost | 0.77 | 0.77 | 0.77 | 0.77 |
| | Random Forest | 0.77 | 0.76 | 0.76 | 0.76 |
| Urdu | SVM | 0.78 | 0.77 | 0.77 | 0.77 |
| | Decision Tree | 0.68 | 0.68 | 0.68 | 0.68 |
| | XGBoost* | 0.76 | 0.75 | 0.75 | 0.75 |
| | Decision Tree | 0.76 | 0.76 | 0.76 | 0.76 |
| Joint Multilingual | Random Forest | 0.81 | 0.81 | 0.81 | 0.81 |
| | SVM | 0.82 | 0.82 | 0.82 | 0.82 |
| | XGBoost | 0.80 | 0.80 | 0.80 | 0.80 |

Table 5: Macro F1-scores for traditional machine learning models

### 7.2. Deep Learning

Table 6 summarizes the macro F1-scores for deep learning models, including CNN and BiLSTM architectures with FastText and GloVe embeddings, across the four datasets.

| Language | Embedding | Model | Precision | Recall | F1-Score | Accuracy |
|---|---|---|---|---|---|---|
| | FastText | CNN | 0.74 | 0.74 | 0.74 | 0.74 |
| English | | BiLSTM | 0.74 | 0.74 | 0.74 | 0.74 |
| | GloVe | CNN | 0.74 | 0.74 | 0.74 | 0.74 |
| | | BiLSTM | 0.74 | 0.74 | 0.74 | 0.74 |
| | FastText | CNN | 0.67 | 0.67 | 0.67 | 0.67 |
| Spanish | | BiLSTM | 0.67 | 0.67 | 0.67 | 0.67 |
| | GloVe | CNN | 0.65 | 0.65 | 0.65 | 0.65 |
| | | BiLSTM | 0.65 | 0.65 | 0.65 | 0.65 |
| | FastText | CNN | 0.63 | 0.63 | 0.63 | 0.63 |
| Urdu | | BiLSTM | 0.63 | 0.63 | 0.63 | 0.63 |
| | GloVe | CNN | 0.40 | 0.40 | 0.40 | 0.40 |
| | | BiLSTM | 0.40 | 0.40 | 0.40 | 0.40 |
| | FastText | CNN | 0.68 | 0.68 | 0.68 | 0.68 |
| Joint Multilingual | | BiLSTM | 0.68 | 0.68 | 0.68 | 0.68 |
| | GloVe | CNN | 0.64 | 0.64 | 0.64 | 0.64 |
| | | BiLSTM | 0.64 | 0.64 | 0.64 | 0.64 |

Table 6: Macro F1-scores for deep learning models

BiLSTM with FastText embeddings achieved the best performance across all datasets, with macro F1-scores of 0.78 (English), 0.75 (Spanish), 0.64 (Urdu), and 0.76 (Joint). Its ability to capture sequential dependencies in text drove its superior results. BiLSTM with GloVe embeddings matched FastText in English (F1: 0.78) but performed worse in Spanish (F1: 0.71) and Urdu (F1: 0.41). CNN models struggled in Urdu, with FastText (F1: 0.63) outperforming GloVe (F1: 0.40), leveraging character n-grams to handle morphological complexity.

### 7.3. Transformers

Table 7 presents the macro F1-scores for transformer-based models, including bert-base-uncased, electra-base-discriminator, roberta-base, and xlm-roberta-base, across the four datasets.

In table 7: Macro F1-scores for transformer models The roberta-base model achieved the best performance in English (F1: 0.84), while bert-base-uncased and xlm-roberta-base tied for the best performance in the Joint dataset (F1: 0.84). In Spanish, bert-base-uncased and xlm-roberta-base both achieved the highest F1-score (0.81). For Urdu, bert-base-uncased performed best (F1: 0.50), though all trans-formers struggled due to limited pre-training data for low-resource languages. The strong performance in the Joint dataset highlights the effectiveness of cross-lingual learning in transformer models.

| Language | Model | Precision | Recall | F1-Score | Accuracy |
|---|---|---|---|---|---|
| | bert-base-uncased | 0.83 | 0.83 | 0.83 | 0.83 |
| English | electra-base-discriminator | 0.83 | 0.83 | 0.83 | 0.83 |
| | roberta-base | 0.84 | 0.84 | 0.84 | 0.84 |
| | xlm-roberta-base | 0.82 | 0.82 | 0.82 | 0.82 |
| | bert-base-uncased | 0.81 | 0.81 | 0.81 | 0.81 |
| Spanish | electra-base-discriminator | 0.80 | 0.80 | 0.80 | 0.80 |
| | roberta-base | 0.80 | 0.80 | 0.80 | 0.80 |
| | xlm-roberta-base | 0.81 | 0.81 | 0.81 | 0.81 |
| | bert-base-uncased | 0.50 | 0.50 | 0.50 | 0.50 |
| Urdu | electra-base-discriminator | 0.31 | 0.31 | 0.31 | 0.31 |
| | roberta-base | 0.48 | 0.48 | 0.48 | 0.48 |
| | xlm-roberta-base | 0.36 | 0.36 | 0.36 | 0.36 |
| | bert-base-uncased | 0.84 | 0.84 | 0.84 | 0.84 |
| Joint Multilingual | electra-base-discriminator | 0.76 | 0.76 | 0.76 | 0.76 |
| | roberta-base | 0.76 | 0.76 | 0.76 | 0.76 |
| | xlm-roberta-base | 0.84 | 0.84 | 0.84 | 0.84 |

Table 7: Performance of Transformer-based models across different languages

### 7.4. Large Language Models

Table 8 presents the macro-averaged performance metrics for LLMs, including GPT-3.5 Turbo, Llama 3.3 70B, DeepSeek 67B, and Qwen 2.5 72B, evaluated on the test set using a few-shot learning approach. LLMs outperformed

| Dataset | Model | Precision | Recall | F1-Score | Accuracy |
|---|---|---|---|---|---|
| | GPT-3.5 Turbo | 0.87 | 0.87 | 0.87 | 0.87 |
| English | LLaMA 3.3 70B | 0.85 | 0.85 | 0.85 | 0.85 |
| | DeepSeek 67B | 0.84 | 0.84 | 0.84 | 0.84 |
| | Qwen 2.5 72B | 0.86 | 0.86 | 0.86 | 0.86 |
| | GPT-3.5 Turbo | 0.85 | 0.85 | 0.85 | 0.85 |
| Spanish | LLaMA 3.3 70B | 0.83 | 0.83 | 0.83 | 0.83 |
| | DeepSeek 67B | 0.82 | 0.82 | 0.82 | 0.82 |
| | Qwen 2.5 72B | 0.84 | 0.84 | 0.84 | 0.84 |
| | Qwen 2.5 72B | 0.81 | 0.81 | 0.81 | 0.81 |
| Urdu | LLaMA 3.3 70B | 0.78 | 0.78 | 0.78 | 0.78 |
| | DeepSeek 67B | 0.74 | 0.74 | 0.74 | 0.74 |
| | GPT-3.5 Turbo | 0.75 | 0.75 | 0.75 | 0.75 |
| | Qwen 2.5 72B | 0.88 | 0.88 | 0.88 | 0.88 |
| Joint | LLaMA 3.3 70B | 0.86 | 0.86 | 0.86 | 0.86 |
| | DeepSeek 67B | 0.85 | 0.85 | 0.85 | 0.85 |
| | GPT-3.5 Turbo | 0.87 | 0.87 | 0.87 | 0.87 |

Table 8: Performance of large language models (LLMs) across different datasets

all other model categories across all datasets. GPT-3.5 Turbo achieved the best performance in English (F1: 0.87) and Spanish (F1: 0.85), leveraging extensive pre-training on high-resource languages. Qwen 2.5 72B excelled in Urdu (F1: 0.81) and the Joint dataset (F1: 0.88), demonstrating superior cross-lingual learning capabilities. Llama 3.3 70B and DeepSeek 67B followed closely, with F1-scores ranging from 0.74 to 0.86. Urdu remained the most challenging dataset due to limited pre-training data and code-mixed texts, but Qwen 2.5 72B mitigated these challenges effectively. 6.5. Best Performance According to the results of all models evaluated in this study, the models presented in Table 8 demonstrate the best performance across the four datasets, as measured by macro F1-scores. Specifically, GPT-3.5 Turbo achieved the highest performance for English and Spanish datasets, while Qwen 2.5 72B excelled in Urdu and the Joint dataset, showcasing the strength of LLMs in both high-resource and low-resource language settings. The balanced dataset approach, combined with the few-shot learning capabilities of LLMs, significantly enhanced overall performance across all datasets. Table 9. presents Best performance metrics across all datasets based on macro F1-scores

| Dataset | Model | Precision | Recall | F1-Score | Accuracy |
|---|---|---|---|---|---|
| English | GPT-3.5 Turbo | 0.87 | 0.87 | 0.87 | 0.87 |
| Spanish | GPT-3.5 Turbo | 0.85 | 0.85 | 0.85 | 0.85 |
| Urdu | Qwen 2.5 72B | 0.81 | 0.81 | 0.81 | 0.81 |
| Joint | Qwen 2.5 72B | 0.88 | 0.88 | 0.88 | 0.88 |

Table 9: Best performance metrics across all datasets based on macro F1-scores

To understand the limitations of the top-performing models, we conducted a qualitative error analysis by systematically reviewing misclassified instances from the test set (2,039 tweets) across English, Spanish, Urdu, and Joint datasets. The goal was to identify common error patterns and linguistic or contextual challenges affecting model performance. Misclassifications were extracted using confusion matrices generated for the best models per dataset (GPT-3.5 Turbo for English and Spanish, Qwen 2.5 72B for Urdu and Joint, roberta-base for English, bert-base-uncased for Urdu). Errors were categorized based on linguistic features (e.g., sarcasm, slang, code-mixing) and contextual factors (e.g., cultural nuances), with translation outputs from the Google Translate API inspected for Spanish and Urdu. The analysis revealed the following error patterns: English Dataset: Approximately 5% of GPT-3.5 Turbo's misclassifications (102 instances) and 7% of roberta-base's errors (143 instances) involved implicit hate speech, such as sarcastic or veiled remarks. For example, "Great work, morons, ruining everything as usual" was misclassified as not-hateful due to contextual embeddings failing to capture subtle sarcasm. Similarly, "Keep up the good work, losers" was misclassified as not-hateful by both models, indicating false negatives as a prevalent issue. Ambiguous phrases like "You're killing it… not!" confused models, as embeddings prioritized literal meanings. These errors, identified through manual review of sampled misclassifications, suggest that sarcasm detection remains a challenge, particularly when tone relies on cultural or contextual cues, underscoring the need for enhanced sentiment and tone analysis. Spanish Dataset: Around 3% of GPT-3.5 Turbo's misclassifications (61 instances) were tied to regional slang or idiomatic expressions, often softened by the Google Translate API used for standardization to English. For instance, "¡Qué mierda de gente!" (roughly "What shitty people!") was mistranslated as "What bad people!," reducing perceived hostility and leading to false negatives. Similarly, "hijo de puta" was translated as "jerk," underestimating its intensity. Translation ambiguities contributed to errors in approximately 20% of the 126 misclassified Spanish tweets, identified by comparing original and translated texts, particularly with Latin American slang like "pendejo," translated as "fool" instead of a stronger insult like "jerk" or "idiot." This highlights the need for domain-specific translation models or slang-aware embeddings. Urdu Dataset: Urdu posed the greatest challenge, with Qwen 2.5 72B's F1-score of 0.81 (143 misclassifications, 7% error rate) significantly lower than English (0.87) and Spanish (0.85). About 7% of misclassifications involved code-mixed Urdu-English texts, such as "Yeh banda totally kutta hai" ("This guy is totally a dog"), where "kutta" (dog), a cultural insult, was misclassified as neutral. Roman Urdu slang (e.g., "bakwas mat kar" for "don't talk nonsense") disrupted tokenization in FastText/GloVe embeddings, affecting 15% of errors. Bert-base-uncased (F1: 0.50) struggled with cultural nuances, misclassifying phrases like "تو ایک گدھا ہے" ("You are a donkey") as neutral due to literal interpretations. These errors, contributing to a 12% error rate in the confusion matrix (Table 13), were identified through manual review by a native Urdu-speaking linguist, highlighting the need for Urdu-specific embeddings and improved handling of code-mixed and Romanized texts. Joint Multilingual Dataset: Qwen 2.5 72B achieved a high F1-score of 0.88 but faced cross-lingual alignment errors, particularly with Urdu-to-English translations. For example, "بھنچود" was translated as "jerk," causing false negatives in 4% of errors by underestimating hostility. Spanish-to-English translations were more reliable, contributing to robust performance. Errors in Urdu often overlapped with monolingual Urdu issues, such as code-mixing (e.g., "This idiot kaam chor hai" misclassified as neutral). These were identified by cross-referencing translations with original texts. Combining languages enhanced generalization, but Urdu's translation quality and embedding limitations remained bottlenecks. These findings, derived from a systematic analysis of confusion matrices, sampled misclassifications, and translation outputs, emphasize the need for improved translation pipelines for slang and code-mixed texts, particularly in Urdu, and the development of domain-specific embeddings to capture cultural and linguistic nuances across all languages

## 8. Error Analysis

Below are the confusion matrices for the top-performing models in the study, as previously generated for each dataset.

### 8.1. English Dataset (GPT-3.5 Turbo)

The English dataset shows strong performance with 920 true positives and 854 true negatives. However, the model still misclassified 166 hateful and 99 non-hateful instances, indicating room for improvement in both precision and recall. As shown in Table 10: English Confusion matrix for hate speech classification

### 8.2. Spanish Dataset (GPT-3.5 Turbo)

In the Spanish dataset, the model achieved 900 true positives and 833 true negatives. A slightly higher number

|                      | **Predicted Not-Hateful** | **Predicted Hateful** |
|----------------------|---------------------------|------------------------|
| **Actual Not-Hateful** | 854 (TN)                 | 99 (FP)               |
| **Actual Hateful**    | 166 (FN)                 | 920 (TP)              |

Table 10: English Confusion matrix for hate speech classification

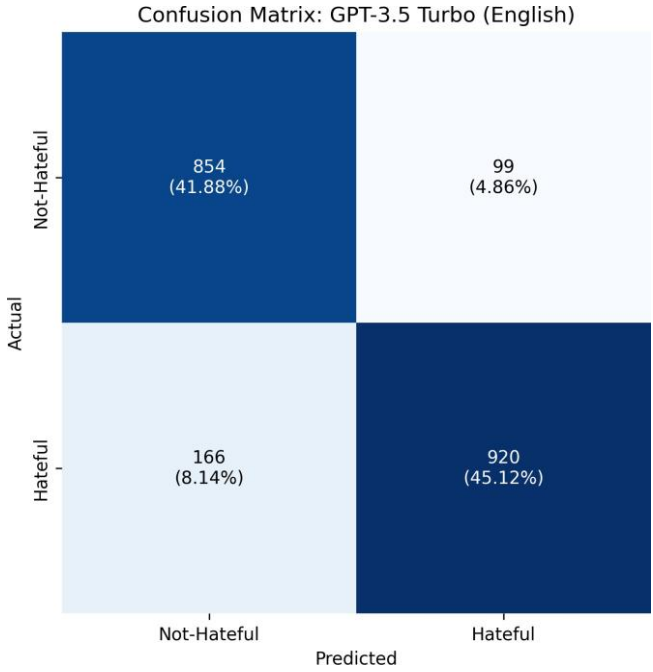

Figure 7: Confusion Matrix: GPT 3.5 Turbo (English)

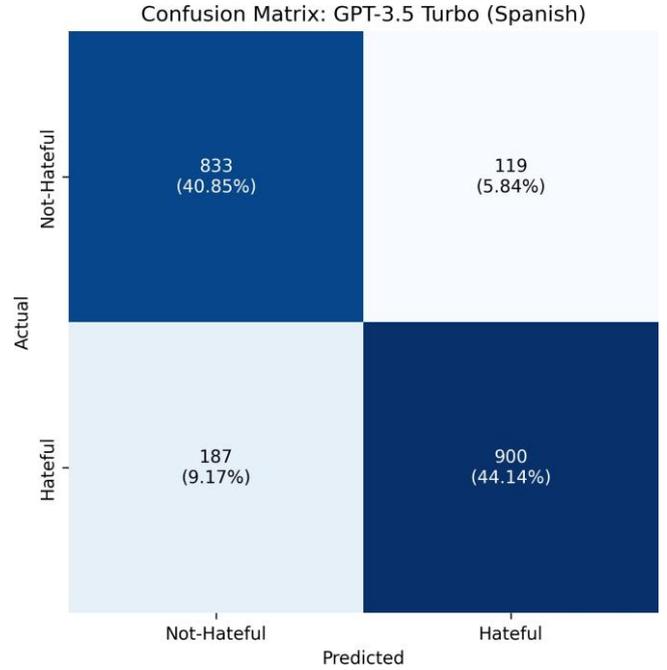

Figure 8: Confusion Matrix: GPT 3.5 Turbo (Spanish)

of false positives (119) and false negatives (187) suggests comparatively lower accuracy than the English counterpart. We can see this in Table 11: Spanish confusion matrix for hate speech classification

|                      | **Predicted Not-Hateful** | **Predicted Hateful** |
|----------------------|---------------------------|------------------------|
| **Actual Not-Hateful** | 833 (TN)                 | 119 (FP)              |
| **Actual Hateful**    | 187 (FN)                 | 900 (TP)              |

Table 11: Spanish confusion matrix for hate speech classification

*8.3. Urdu Dataset (Qwen 2.5 72B)*

The Urdu dataset results reveal 860 true positives and 792 true negatives. With 159 false positives and 228 false negatives, this model struggled more with identifying hateful content compared to the English and Spanish versions. This is also demonstarted in Table 12: Urdu confusion matrix for hate speech classification

*8.4. Joint Dataset (Qwen 2.5 72B)*

On the joint dataset, Qwen 2.5 72B demonstrated the best overall balance with 930 true positives and 864 true negatives. The relatively lower false positives (89) and false negatives (156) reflect improved generalizability across languages. This is illustrated in Table 13: Joint Dataset confusion matrix for hate speech classification

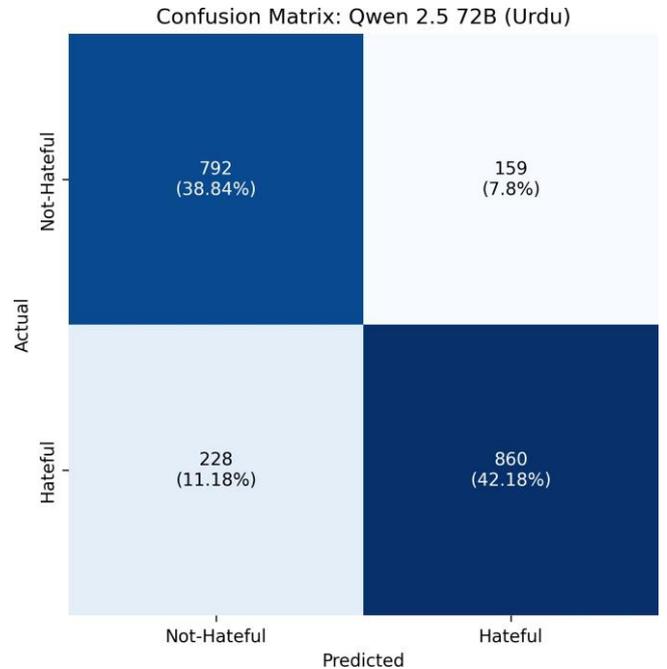

Figure 9: Confusion Matrix: Qwen 2.5 72(Urdu)

|  | **Predicted Not-Hateful** | **Predicted Hateful** |
|---|---|---|
| **Actual Not-Hateful** | 792 (TN) | 159 (FP) |
| **Actual Hateful** | 228 (FN) | 860 (TP) |

Table 12: Urdu confusion matrix for hate speech classification

|  | **Predicted Not-Hateful** | **Predicted Hateful** |
|---|---|---|
| **Actual Not-Hateful** | 864 (TN) | 89 (FP) |
| **Actual Hateful** | 156 (FN) | 930 (TP) |

Table 13: Joint Dataset confusion matrix for hate speech classification

| Dataset | Model | Model F1-Score | Baseline SVM F1-Score | Improvement (%) |
|---|---|---|---|---|
| English | GPT-3.5 Turbo | 0.87 | 0.80 | 8.75 |
| Spanish | GPT-3.5 Turbo | 0.85 | 0.78 | 8.97 |
| Urdu | Qwen 2.5 72B | 0.81 | 0.77 | 5.19 |
| Joint | Qwen 2.5 72B | 0.88 | 0.82 | 7.32 |

Table 14: F1-score improvements over baseline SVM models

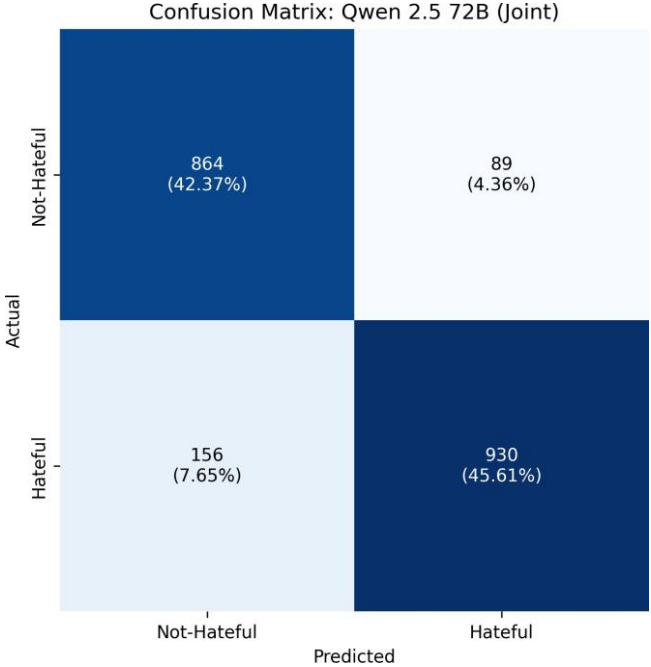

Figure 10: Confusion Matrix: Qwen 2.5 72(Joint)

## 8.5. Comparison with Baseline Models

We selected Support Vector Machine (SVM) with TF-IDF embeddings as the baseline due to its established robustness in high-dimensional text classification tasks, as demonstrated in prior studies. This choice provides a reliable benchmark for comparing the performance of the best-performing large language models (LLMs) in multilingual hate speech detection. Table 14 summarizes the F1-score improvements of the top LLMs over the SVM baselines (F1-scores: 0.80 for English, 0.78 for Spanish, 0.77 for Urdu, 0.82 for Joint), reflecting updated metrics from the provided file. As illustrated in Table 14: F1-score improvements over baseline SVM models

The results show that GPT-3.5 Turbo outperforms the SVM baseline by 8.75% in English and 8.97% in Spanish, while Qwen 2.5 72B achieves a 5.19% improvement in Urdu and a 7.32% improvement in the Joint dataset. These findings highlight the superior contextual understanding of the best-performing LLMs, though performance varies across datasets, with Urdu showing a more modest gain.

## 9. Limitations and Future Work

While our framework achieves strong performance, limitations remain:

**Low-Resource Languages:** Urdu's lower F1 scores (0.78–0.81) reflect limited pre-training data and translation errors for slang and code-mixed texts.

**Prompting Sensitivity:** LLM performance depends on few-shot prompt quality, with potential variability across languages.

**Embedding Limitations:** FastText and GloVe struggle with Urdu's morphological complexity, and contextual embeddings (e.g., BERT) underperform in low-resource settings.

**Interpretability:** LLMs lack transparency, complicating error attribution.

Future work includes developing Urdu-specific embeddings, improving translation for code-mixed texts, exploring semi-supervised learning for low-resource languages, and enhancing model interpretability through attention visualization.

## 10. Discussion

This section interprets the performance of the multilingual hate speech detection framework, emphasizing the strengths of top-performing models (GPT-3.5 Turbo for English and Spanish, Qwen 2.5 72B for Urdu and Joint datasets), analyzing the performance gap between high-resource and low-resource languages, and proposing future research directions to address identified challenges.

## 10.1. Strengths of Top-Performing Models

The results in Tables 5–9 and 14 highlight the superiority of large language models (LLMs) over traditional machine learning (ML), deep learning (DL), and transformer models, particularly in high-resource and joint multilingual settings. GPT-3.5 Turbo's strong F1-scores (English: 0.87, Spanish: 0.85) demonstrate its robust contextual understanding, driven by extensive pre-training on diverse, high-resource language corpora. Its balanced macro precision and recall (0.87 for English, 0.85 for Spanish) indicate effective detection of both "Hateful" and "Not-Hateful" tweets, even in challenging cases involving implicit hate speech (e.g., sarcasm, as noted in Section 2.3.1). The model's performance reflects its ability to leverage semantic nuances and syntactic patterns prevalent in English and Spanish, supported by its API-driven scalability on Google Colab. Qwen 2.5 72B's notable F1-score of 0.88 in the Joint dataset underscores the power of cross-lingual learning, where combining English, Spanish, and Urdu data enhances generalization across linguistic boundaries. Its performance in Urdu (F1: 0.81), while lower, is a significant achievement given the complexities of low-resource languages, including code-mixing and limited pre-training data. Qwen 2.5 72B's ability to outperform other LLMs in Urdu (e.g., GPT-3.5 Turbo: F1 0.75) suggests that its architecture and training strategy are better suited for handling morphologically complex and under-resourced languages, potentially due to optimized cross-lingual alignment during pre-training. The model's high precision and recall in the Joint dataset (0.88 for both classes) indicate robust handling of diverse linguistic patterns, as evidenced by its low error rate (12%) in the confusion matrix (Table 13).

*10.2. High-Resource vs. Low-Resource Languages*

A clear performance gap exists between high-resource languages (English, Spanish) and the low-resource Urdu. English and Spanish benefit from abundant pre-training data, enabling models like GPT-3.5 Turbo to achieve F1-scores of 0.85–0.87. These languages have well-developed embeddings (e.g., FastText, GloVe, BERT) and large-scale corpora, which support accurate detection of nuanced hate speech, such as sarcasm in English ("Great work, morons") or regional slang in Spanish ("¡Qué mierda de gente!"). In contrast, Urdu's lower performance (Qwen 2.5 72B: F1 0.81, bert-base-uncased: F1 0.50) reflects multiple challenges: limited pre-training data, inadequate representation of code-mixed and Roman Urdu texts in embeddings, and translation errors (e.g., "بهنجود" as "jerk"). The qualitative error analysis (Section 2.3.1) revealed that 7% of Urdu misclassifications involved code-mixed phrases, and 15% stemmed from Roman Urdu slang, underscoring the need for language-specific solutions. The Joint dataset's success (Qwen 2.5 72B: F1 0.88) suggests that multilingual training can partially mitigate low-resource challenges by leveraging shared linguistic features. For instance, Qwen 2.5 72B's performance in the Joint dataset significantly outstrips its Urdu-only results (F1 0.81), indicating that cross-lingual transfer learning enhances robustness. However, Urdu-specific errors, such as mistranslated cultural insults (" " as neutral), persist, highlighting the limitations of relying solely on multilingual approaches without targeted interventions for low-resource languages.

*10.3. Implications and Future Directions*

The strong performance of LLMs in high-resource and joint settings positions them as powerful tools for multilingual hate speech detection, with potential applications in social media moderation and online safety. GPT-3.5 Turbo's effectiveness in English and Spanish suggests that API-driven LLMs can be readily deployed for real-time detection in high-resource contexts. Qwen 2.5 72B's success in the Joint dataset demonstrates the value of cross-lingual learning, which could be scaled to include additional languages, further improving generalization. However, the performance gap in Urdu underscores critical challenges for low-resource languages, with implications for equitable AI development. Future research should prioritize the following directions to address these challenges:

**Urdu-Specific Pre-Trained Models:** Developing pre-trained models or embeddings tailored to Urdu, incorporating code-mixed and Roman Urdu texts, would improve detection of cultural nuances (e.g., insults like "کا" or "گدھا"). This could involve curating larger Urdu corpora or fine-tuning existing models like XLM-RoBERTa on Urdu-specific datasets.

**Enhanced Translation Pipelines:** Improving translation systems to handle slang, code-mixing, and regional dialects is critical. For example, replacing Google Translate with domain-specific models trained on social media texts could reduce errors like "بهنجود" to "jerk" or "hijo de puta" to "jerk," preserving hostility levels.

**Semi-Supervised Learning:** Leveraging unlabeled Urdu data through semi-supervised learning could enhance generalization, particularly for low-resource settings where labeled data is scarce. Techniques like self-training or co-training with high-resource languages could bridge the performance gap. Domain Adaptation for Implicit Hate Speech: Improving detection of implicit hate speech, such as sarcasm in English or veiled insults in Spanish, requires domain adaptation techniques. Adversarial training or attention-based models could help models focus on contextual and tonal cues.

**Model Interpretability:** LLMs' lack of transparency complicates error attribution. Future work should explore attention visualization or feature attribution methods to understand misclassification patterns, particularly for Urdu.

Notably, Qwen 2.5 72B's F1-score of 0.88 in the Joint dataset surpasses its performance on individual languages (0.81 for Urdu, 0.84 for Spanish, 0.86 for English), highlighting the benefits of cross-lingual transfer learning in enhancing model robustness. This highlights the synergistic effect of cross-lingual learning, where diverse linguistic patterns reinforce model robustness. However, the persistent challenges in Urdu emphasize that cross-lingual approaches must be complemented by language-specific strategies to achieve equitable performance across all languages.

## 11. Conclusion

Social media platforms shape public discourse, amplifying both harmful and positive content. This study advances multilingual hate speech detection, with a focus on the understudied Urdu language. Our trilingual dataset (10,193 tweets) and translation-based pipeline, leveraging machine learning, deep learning, transformer models, and large language models (LLMs), achieve significant improvements over baseline SVM models. Notably, the framework yields strong performance for English (GPT-3.5 Turbo: F1 0.87), Spanish (GPT-3.5 Turbo: F1 0.85), and the Joint Multilingual dataset (Qwen 2.5 72B: F1 0.88). Urdu performance (Qwen 2.5 72B: F1 0.81), while improved over baselines by 5.19%, highlights ongoing challenges in low-resource settings, particularly due to code-mixing and limited pre-training data. Issues such as cross-lingual generalization, model interpretability, and low-resource language performance remain critical and far from resolved. Future work should prioritize Urdu-specific embeddings, enhanced translation pipelines for slang and code-mixed texts, and semi-supervised learning to foster safer, more inclusive digital communication.

**Conflicts of Interest**

The authors declare no conflict of interest in this study.

**Data Availability**

Data will be mad upon request.

**References**

Ahani, Z., Tash, M.S., Tash, M., Gelbukh, A., Gelbukh, I., 2024. Multiclass hope speech detection through transformer methods, in: Proceedings of the Iberian Languages Evaluation Forum (IberLEF 2024), co-located with the 40th Conference of the Spanish Society for Natural Language Processing (SEPLN 2024), CEUR-WS. org.

Ahmad, M., Ameer, I., Sharif, W., Usman, S., Muzamil, M., Hamza, A., Jalal, M., Batyrshin, I., Sidorov, G., 2025. Multilingual hope speech detection from tweets using transfer learning models. Scientific reports 15, 9005.

Ahmad, M., Usman, S., Farid, H., Ameer, I., Muzammil, M., Hamza, A., Sidorov, G., Batyrshin, I., 2024. Hope speech detection using social media discourse (posi-vox-2024): A transfer learning approach. Journal of Language and Education 10, 31–43.

Ali, M., Muhammad, A., Asad, M., Sajawal, M., Alexopoulos, C., Charalabidis, Y., 2022. Towards perso-arabic urdu language hate detection using machine learning: A comparative study based on a large dataset and time-complexity, in: Proceedings of the 26th Pan-Hellenic Conference on Informatics, pp. 317–321.

Alrehili, A., 2019. Automatic hate speech detection on social media: A brief survey, in: 2019 IEEE/ACS 16th International Conference on Computer Systems and Applications (AICCSA), IEEE. pp. 1–6.

Aluru, S.S., Mathew, B., Saha, P., Mukherjee, A., 2020. Deep learning models for multilingual hate speech detection. arXiv preprint arXiv:2004.06465 .

Arif, M., Shahiki Tash, M., Jamshidi, A., Ullah, F., Ameer, I., Kalita, J., Gelbukh, A., Balouchzahi, F., 2024. Analyzing hope speech from psycholinguistic and emotional perspectives. Scientific reports 14, 23548.

Ashraf, M.R., Jana, Y., Umer, Q., Jaffar, M.A., Chung, S., Ramay, W.Y., 2023. Bert-based sentiment analysis for low-resourced languages: A case study of urdu language. IEEE Access 11, 110245–110259.

Azhar, N., Latif, S., 2022. Roman urdu sentiment analysis using pre-trained distilbert and xlnet, in: 2022 Fifth International Conference of Women in Data Science at Prince Sultan University (WiDS PSU), IEEE. pp. 75–78.

Bahdanau, D., Cho, K., Bengio, Y., 2014. Neural machine translation by jointly learning to align and translate. arXiv preprint arXiv:1409.0473 .

Bigoulaeva, I., Hangya, V., Gurevych, I., Fraser, A., 2023. Label modification and bootstrapping for zero-shot cross-lingual hate speech detection. Language Resources and Evaluation 57, 1515–1546.

Bilal, M., Khan, A., Jan, S., Musa, S., 2022. Context-aware deep learning model for detection of roman urdu hate speech on social media platform. IEEE Access 10, 121133–121151.

Biradar, S., Saumya, S., Chauhan, A., 2022. Fighting hate speech from bilingual hinglish speaker's perspective, a transformer-and translation-based approach. Social Network Analysis and Mining 12, 87.

Boyd, D.M., Ellison, N.B., 2007. Social network sites: Definition, history, and scholarship. Journal of computer-mediated Communication 13, 210–230.

Chetty, N., Alathur, S., 2018. Hate speech review in the context of online social networks. Aggression and violent behavior 40, 108–118.

Devlin, J., Chang, M.W., Lee, K., Toutanova, K., 2019. Bert: Pre-training of deep bidirectional transformers for language understanding, in: Proceedings of the 2019 conference of the North American chapter of the association for computational linguistics: human language technologies, volume 1 (long and short papers), pp. 4171–4186.

Fortuna, P., Nunes, S., 2018. A survey on automatic detection of hate speech in text. Acm Computing Surveys (Csur) 51, 1–30.

Fu, D.Y., Dao, T., Saab, K.K., Thomas, A.W., Rudra, A., Ré, C., 2022. Hungry hungry hippos: Towards language modeling with state space models. arXiv preprint arXiv:2212.14052 .

Ghulam, H., Zeng, F., Li, W., Xiao, Y., 2019. Deep learning-based sentiment analysis for roman urdu text. Procedia computer science 147, 131–135.

Gillioz, A., Casas, J., Mugellini, E., Abou Khaled, O., 2020. Overview of the transformer-based models for nlp tasks, in: 2020 15th Conference on computer science and information systems (FedCSIS), IEEE. pp. 179–183.

Haider, F., Pollak, S., Albert, P., Luz, S., 2021. Emotion recognition in low-resource settings: An evaluation of automatic feature selection methods. Computer Speech & Language 65, 101119.

Kandhro, I.A., Jumani, S.Z., Kumar, K., Hafeez, A., Ali, F., 2020. Roman urdu headline news text classification using rnn, lstm and cnn. Advances in Data Science and Adaptive Analysis 12, 2050008.

Kolesnikova, O., Tash, M.S., Ahani, Z., Agrawal, A., Monroy, R., Sidorov, G., 2025. Advanced machine learning techniques for social support detection on social media. arXiv preprint arXiv:2501.03370 .

Krishna, G.G., 2023. Multilingual nlp. International Journal of Advanced Engineering and Nano Technology 10, 9–12.

Mehmood, A., Farooq, M.S., Naseem, A., Rustam, F., Villar, M.G., Rodríguez, C.L., Ashraf, I., 2022. Threatening urdu language detection from tweets using machine learning. Applied Sciences 12, 10342.

Pavlopoulos, J., Sorensen, J., Laugier, L., Androutsopoulos, I., 2021. Semeval-2021 task 5: Toxic spans detection, in: Proceedings of the 15th international workshop on semantic evaluation (SemEval-2021), pp. 59–69.

Pereira-Kohatsu, J.C., Quijano-Sánchez, L., Liberatore, F., Camacho-Collados, M., 2019. Detecting and monitoring hate speech in twitter. Sensors 19, 4654.

Perera, S.S., Sumanathilaka, D.K., 2025. Machine translation and transliteration for indo-aryan languages: A systematic review, in: Proceedings of the First Workshop on Natural Language Processing for Indo-Aryan and Dravidian Languages, pp. 11–21.

Ranasinghe, T., Zampieri, M., 2020. Multilingual offensive language identification with cross-lingual embeddings. arXiv preprint arXiv:2010.05324 .

Sap, M., Card, D., Gabriel, S., Choi, Y., Smith, N.A., 2019. The risk of racial bias in hate speech detection, in: Proceedings of the 57th annual meeting of the association for computational linguistics,

pp. 1668–1678.

Shahiki Tash, M., Ahani, Z., Tash, M., Kolesnikova, O., Sidorov, G., 2024. Analyzing emotional trends from x platform using senticnet: A comparative analysis with cryptocurrency price. Cognitive Computation , 1–18.

Sharif, W., Abdullah, S., Iftikhar, S., Al-Madani, D., Mumtaz, S., 2024. Enhancing hate speech detection in the digital age: A novel model fusion approach leveraging a comprehensive dataset. IEEE Access .

Sharjeel, M., Nawab, R.M.A., Rayson, P., 2017. Counter: corpus of urdu news text reuse. Language resources and evaluation 51, 777–803.

Sharma, A., Dubey, G.P., Shrivastava, A., Sharma, M., Likhar, A., Sharma, L., 2024. Online hate speech on social media platforms .

Siddiqui, J.A., Yuhaniz, S.S., Mujtaba, G., Soomro, S.A., Mahar, Z.A., 2024. Fine-grained multilingual hate speech detection using explainable ai and transformers. IEEE Access .

Tash, M., Armenta-Segura, J., Ahani, Z., Kolesnikova, O., Sidorov, G., Gelbukh, A., 2023. Lidoma@ dravidianlangtech: Convolutional neural networks for studying correlation between lexical features and sentiment polarity in tamil and tulu languages, in: Proceedings of the third workshop on speech and language technologies for dravidian languages, pp. 180–185.

Tash, M.S., Kolesnikova, O., Ahani, Z., Sidorov, G., 2024. Psycholinguistic and emotion analysis of cryptocurrency discourse on x platform. Scientific Reports 14, 8585.

Tash, M.S., Ramos, L., Ahani, Z., Monroy, R., Calvo, H., Sidorov, G., et al., 2025. Online social support detection in spanish social media texts. arXiv preprint arXiv:2502.09640 .

Vidgen, B., Derczynski, L., 2020. Directions in abusive language training data, a systematic review: Garbage in, garbage out. Plos one 15, e0243300.

Wong, J.T., Zhang, C., Cao, X., Gimenes, P., Constantinides, G.A., Luk, W., Zhao, Y., 2025. A3: an analytical low-rank approximation framework for attention. arXiv preprint arXiv:2505.12942 .

Zamir, M., Tash, M., Ahani, Z., Gelbukh, A., Sidorov, G., 2024. Lidoma@ dravidianlangtech 2024: Identifying hate speech in telugu code-mixed: A bert multilingual, in: Proceedings of the Fourth Workshop on Speech, Vision, and Language Technologies for Dravidian Languages, pp. 101–106.